\pdfoutput=1

\documentclass[11pt]{article}
\usepackage{acl}

\usepackage{times}
\usepackage{latexsym}

\usepackage[T1]{fontenc}
\usepackage[utf8]{inputenc}

\usepackage{microtype}

\PassOptionsToPackage{hyphens}{url}
\usepackage{hyperref}
\usepackage[english]{babel}
\usepackage{hyphenat}
\usepackage{natbib}
\usepackage{amsmath}
\usepackage{amssymb}
\usepackage{amsfonts}
\usepackage{bm}
\usepackage{booktabs}
\usepackage{graphicx}
\usepackage{subcaption}
\usepackage{mleftright}
\usepackage[detect-weight=true,binary-units=true]{siunitx}
\usepackage[capitalize,nameinlink]{cleveref}
\usepackage{nameref}
\usepackage[export]{adjustbox}
\usepackage{mathtools}
\usepackage{placeins}
\usepackage{csquotes}
\usepackage{todonotes}
\usepackage{paralist}
\usepackage{xcolor}
\usepackage[normalem]{ulem}

\title{Black-box language model explanation\\by context length probing}
\author{\llap{Ondřej Cífka}\quad\rlap{Antoine Liutkus}\\
  Zenith Team, LIRMM, CNRS UMR 5506 \\
  Inria, Université de Montpellier, France \\
  \texttt{cifka@matfyz.cz}, \texttt{antoine.liutkus@inria.fr}
}

\def\huggingface{\raisebox{-0.55ex}{\includegraphics[width=1.3em]{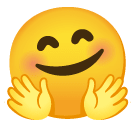}}}

\crefformat{footnote}{#2\footnotemark[#1]#3}

\makeatletter
\newif\iffinalcopy
\ifacl@finalcopy
\finalcopytrue
\fi
\makeatother

\newcommand{\anon}[1]{\iffinalcopy{#1}\else[anonymized]\fi}

\begin{document}

\maketitle

\begin{abstract}
The increasingly widespread adoption of large
language models has highlighted the need for improving their explainability.
We present \emph{context length probing}, a novel explanation technique for causal language models, based on tracking the predictions of a model as a function of the length of available context, and %
allowing to assign \emph{differential importance scores} to different contexts.
The technique is model-agnostic and does not rely on access to model internals beyond computing token-level probabilities.
We apply context length probing to large pre-trained language models and offer some initial analyses and insights, including the potential for studying long-range dependencies.
The source code\footnote{\iffinalcopy\url{https://github.com/cifkao/context-probing/}\else\url{https://anonymous.4open.science/r/context-probing-DBEB}\fi} and an interactive
demo\footnote{\iffinalcopy\url{https://cifkao.github.io/context-probing/}\else\url{https://context-probing.netlify.app/}\fi\label{foot:demo-url}} of the method are available.
\end{abstract}

\section{Introduction}
Large language models (LMs), typically based on the Transformer architecture \cite{VaswaniSPUJGKP17}, have recently seen increasingly widespread adoption, yet understanding their behaviour remains a difficult challenge and an active research topic.

Notably, as the length of the context that can be accessed by LMs has grown, a question that has attracted some attention is how this influences their predictions. Some recent studies in this line of research suggest that even ``long-range'' LMs focus heavily on local context and largely fail to exploit distant ones
\citep{oconnor-andreas-2021-context,sun-etal-2021-long,press-etal-2021-shortformer,sun-etal-2022-chapterbreak}.
A more nuanced understanding of how contexts of different lengths influence LMs' predictions may hence be valuable for further improving their performance, especially on tasks like long-form text generation where long-range dependencies are of critical importance.

\begin{figure}
    \centering
    \includegraphics[width=\linewidth]{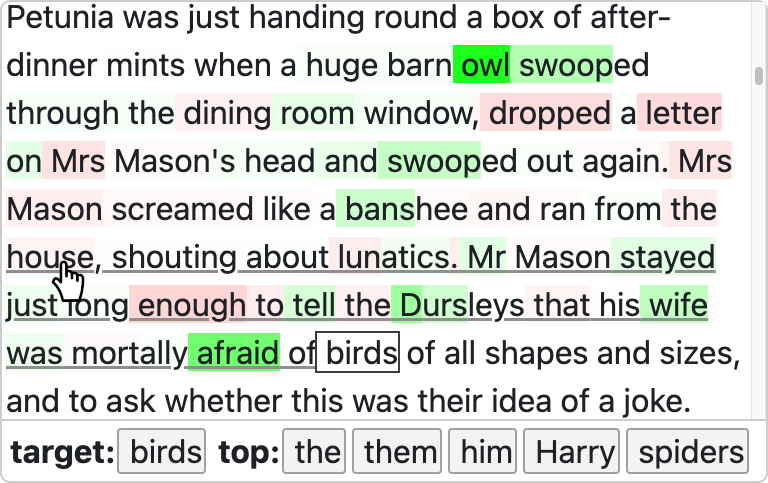}
    \caption{A screenshot of a %
    demo\cref{foot:demo-url} of the proposed method.
    After selecting a target token (here ``\textbf{birds}''), the preceding tokens are highlighted according to their (normalized) \emph{differential importance scores} (green = positive, red = negative), obtained using our method.
    The user can also explore the top predictions for contexts of different lengths (here the context ``house, shouting about lunatics. \textelp{} mortally afraid of'').}
    \label{fig:ui-screenshot}
\end{figure}

In this work, we propose \emph{context length probing}, a simple explanation technique for \emph{causal} (autoregressive) language models, based on tracking the predictions of the model as a function of the number of tokens available as context.
Our proposal has the following advantages:

\begin{itemize}
    \item It is conceptually simple, providing a straightforward answer to a natural question: \emph{How does the length of available context impact the prediction?}
    \item It can be applied to a pre-trained model without retraining or fine-tuning and without training any auxiliary models.
    \item It does not require access to model weights, internal representations or gradients.
    \item It is model-agnostic, as it can be applied to any causal LM, including at\-ten\-tion\-less architectures like RNN \citep{Mikolov2010} and CNN \citep{pmlr-v70-dauphin17a}. The only requirement for the model is to accept arbitrary input segments %
    (i.e.\ not be limited to document prefixes).
\end{itemize}
Furthemore, we propose a way to use this technique %
to assign what we call \emph{differential importance scores} to contexts of different lengths.
This can be seen as complementary to other techniques like attention or saliency map visualization. %
Interestingly, contrary to those techniques, ours appears promising as a tool for studying long-range dependencies, since it can be expected to highlight important information not already covered by shorter contexts.

\section{Related work}
A popular way to dissect Transformers is by visualizing their attention weights \cite[e.g.][]{vig-2019-multiscale,hoover-etal-2020-exbert}.
However, it has been argued that this does not provide reliable explanations and can be misleading \cite{jain-wallace-2019-attention,serrano-smith-2019-attention}.
A more recent line of work \cite{elhage2021mathematical,olsson2022context} explores ``mechanistic explanations'', based on reverse-engineering the computations performed by Transformers.
These techniques are tied to concrete architectures, which are often ``toy'' versions of those used in real-world applications, e.g.\ attention-only Transformers in \citeauthor{elhage2021mathematical} 

Other options include general-purpose methods like neuron/activation interpretation \cite[e.g.][]{geva-etal-2021-transformer,goh2021multimodal,dai-etal-2022-knowledge}, saliency maps \cite[e.g.][]{Fong17Interpretable,pmlr-v97-ancona19a} and influence functions \cite{pmlr-v70-koh17a}.
These require access to internal representations and/or the ability to backpropagate gradients, and have some caveats of their own \cite{DBLP:series/lncs/KindermansHAASDEK19,kokhlikyan2021investigating}. %

More closely related to our work are studies that perform \emph{ablation} (e.g.\ by shuffling, truncation or masking) on different contexts to understand their influence on predictions \citep{oconnor-andreas-2021-context,sun-etal-2021-long,press-etal-2021-shortformer,vafa-etal-2021-rationales}.
To our knowledge, all such existing works only test a few select contexts or greedily search for the most informative one; in contrast, we show that it is feasible to consider \emph{all} context lengths in the range from 1 to a maximum $c_\text{max}$, which permits us to obtain fine-grained insights on the example level, e.g.\ in the form of the proposed differential importance scores.
Moreover, many existing analyses
\citep[e.g.][]{vafa-etal-2021-rationales,oconnor-andreas-2021-context}
rely on specific training or fine-tuning, which is not the case with our proposal.

\section{Method}
\subsection{Context length probing}
A causal LM estimates the conditional probability distribution of a token given its left-hand context in a document:
\begin{equation}
    p\mleft(x_{n+1}\;\middle|\;x_1,\ldots,x_n\mright).
\end{equation}
We are interested here in computing the probabilities conditioned on a \emph{reduced} context of length $c\in\left\{1,\ldots,n\right\}$:
\begin{equation}
    p\mleft(x_{n+1}\;\middle|\;x_{n-c+1},\ldots,x_n\mright),
    \label{eq:reduced-ctx-prob}
\end{equation}
so that we may then study the behavior of this distribution as a function of $c$.

An apparent obstacle in doing so is that applying the model to an arbitrary subsequence $x_{n-c+1},\ldots,x_n$, instead of the full document $x_1,\ldots,x_N$, may lead to inaccurate estimates of the probabilities in \cref{eq:reduced-ctx-prob}.
However, we note that large LMs are not usually trained on entire documents.
Instead, the training data is pre-processed by shuffling all the documents, concatenating them (with a special token as a separator), and splitting the resulting sequence into \emph{chunks} of a fixed length (usually \num{1024} or \num{2048} tokens) with no particular relation to the document length.
Thus, the models are effectively trained to accept sequences of tokens starting at arbitrary positions in a document and it is therefore %
correct to employ them as such to compute estimates of \cref{eq:reduced-ctx-prob}.\footnote{For models trained on data that is pre-processed differently, (re)training or fine-tuning with data augmentation such as random shifts may be needed in order to apply our method, analogously to \citet{vafa-etal-2021-rationales}, who use word dropout to ensure compatibility with their method.}

It now remains to be detailed how to efficiently evaluate the above probabilities for all positions $n$ and context lengths $c$.
Specifically, for a given document $x_1,\ldots,x_N$ and some maximum context length $c_\text{max}$, we are interested in an $\left(N-1\right)\times c_\text{max}\times\lvert\mathcal{V}\rvert$ tensor $\bm{P}$, where $\mathcal{V}=\left\{w_1,\ldots,w_{\lvert\mathcal{V}\rvert}\right\}$ is the vocabulary, such that:
\begin{equation}
    \bm{P}_{n,c,i} = p\mleft(x_{n+1}=w_i\;\middle|\;x_{n-c+1},\ldots,x_n\mright),\label{eq:p}
\end{equation}
with $\bm{P}_{n,c,*}=\bm{P}_{n,n-1,*}$ for $n\leq c$.\footnote{$\bm{P}_{n,c,*}$ is a $\lvert\mathcal{V}\rvert$-dimensional slice of $\bm{P}$ along the last axis.}
Observe that by running the model on any segment $x_m,\ldots,x_n$, we obtain all the values $\bm{P}_{m+c-1,c,*}$ for $c\in\left\{1,\ldots,n-m+1\right\}$.
Therefore, we can fill in the tensor $\bm{P}$ by applying the model along a sliding window of size $c_\text{max}$, i.e.\ running it on $N$ (overlapping) segments of length at most $c_\text{max}$.
See \cref{sec:appendix-context-probing} for an illustration and additional remarks.%

\subsection{Metrics}
\label{sec:metrics}
Having obtained the tensor $\bm{P}$ as we have just described, we use it to study how the predictions evolve as the context length is increased from \num{1} to $c_\text{max}$.
Specifically, our goal is to define a suitable metric that we can compute from $\bm{P}_{n,c,*}$ and follow it as a function of $c$ (for a specific $n$ or on average).

One possibility would be to use the negative log-likelihood (NLL) loss values: \begin{equation}
-\log p\mleft(x_{n+1}\;\middle|\;x_{n-c+1},\ldots,x_n\mright).
\end{equation}
However, this may not be a particularly suitable metric for explainability purposes, as it depends (only) on the probability assigned to the \emph{ground truth} $x_{n+1}$, %
while the LM outputs a probability distribution $\bm{P}_{n,c,*}$ over the entire vocabulary, which may in fact contain many other plausible continuations.
For this reason, we propose to
exploit a metric defined on whole \textit{distributions}, e.g.\ the Kullback-Leibler (KL) divergence. To achieve this, we choose the maximum-context predictions $\bm{P}_{n,c_\text{max},*}$ as a reference and get:
\begin{equation}
\begin{aligned}
    \mathcal{D}_{n,c} &= D_\text{KL}\mleft[\bm{P}_{n,c_\text{max},*}\;\middle\|\;\bm{P}_{n,c,*}\mright]\\
    &= \sum_{i=1}^{\lvert\mathcal{V}\rvert} \bm{P}_{n,c_\text{max},i} \log 
    \frac{\bm{P}_{n,c_\text{max},i}}{\bm{P}_{n,c,i}}.
\end{aligned}
\label{eq:kl-div-metric}
\end{equation}
The rationale for \eqref{eq:kl-div-metric} is to quantify the amount of information that is lost by using a shorter context $c\leq c_\text{max}$. Interestingly, this metric is \textit{not} related to the absolute performance of the model with maximal context, but rather to how the output \textit{changes} if a shorter context is used. %

\subsection{Differential importance scores}
\label{sec:delta-scores}
We are also interested in studying how individual \emph{increments} in context length affect the predictions.
We propose to quantify this as the change in the KL divergence metric \eqref{eq:kl-div-metric} when a new token is introduced into the context.
Specifically, for a pair of tokens $x_{n+1}$ (the \emph{target token}) and $x_m$ (the \emph{context token}), we define a \emph{differential importance score} ($\Delta$-score for short)
\begin{equation}
    \Delta\mathcal{D}_{n, m} = \mathcal{D}_{n,n-m-1}-\mathcal{D}_{n,n-m}.
\end{equation}
We may visualize these scores as a way to explain the LM predictions, much like is often done with attention weights, with two important differences. %
First, a high $\Delta\mathcal{D}_{n,m}$ should not be interpreted as meaning that $x_m$ in isolation is important for predicting $x_{n+1}$, but rather that it is salient given the context that follows it (which might mean that it brings information not contained in the following context).
Second, unlike attention weights, our scores need not sum up to one, and can be negative;
in this regard,
the proposed representation is more conceptually similar to a saliency map than to an attention map.

\section{Results}
We apply the proposed technique to publicly available pre-trained large Transformer language models, namely GPT-J \citep{gpt-j} and two GPT-2 \citep{radford2019language} variants~-- see \cref{tab:models} for an overview.
We use the validation set of the English LinES treebank\footnote{\url{https://universaldependencies.org/treebanks/en_lines/index.html}} from Universal Dependencies (UD; \citealp{nivre-etal-2020-universal}), containing 8 documents with a total length of \num{20672} tokens\footnote{After concatenating all sentences and applying the GPT-2 tokenizer, which is used by both GPT-2 and GPT-J.} and covering fiction, an online manual, and Europarl data.
We set $c_\text{max}=1023$.
We use the \huggingface~Transformers library\footnote{\url{https://github.com/huggingface/transformers}}\ \citep{wolf-etal-2020-transformers-img} to load the pre-trained models and run inference.
Further technical details are included in \cref{sec:appendix-computation-details}.

\begin{table}
    \centering
    \begin{tabular}{l@{\;\,}r@{\,}l@{\;\,}r@{\;\,}r@{\;\,}r@{\;\,}r@{\;\,}r}
        \toprule
        {name} & \multicolumn{2}{@{}c@{\;\,}}{\hskip-1mm\#param} & {\#layer} & {\#head} & {$d_\text{model}$} & {max len} \\
        \midrule
        gpt2 & 117 & M & 12 & 12 & 768 & 1024 \\
        gpt2-xl & 1.5 & B & 48 & 25 & 1600 & 1024 \\
        gpt-j-6B & 6.1 & B & 28 & 16 & 4096 & 2048 \\
        \bottomrule
    \end{tabular}
    \caption{Hyperparameters of the 3 models used.}
    \label{tab:models}
\end{table}

\subsection{LM loss by context length}
\label{sec:loss-by-context-length}

\begin{figure}
    \centering
    \includegraphics[width=0.95\linewidth,trim={0 0.25cm 0 0.2cm},clip]{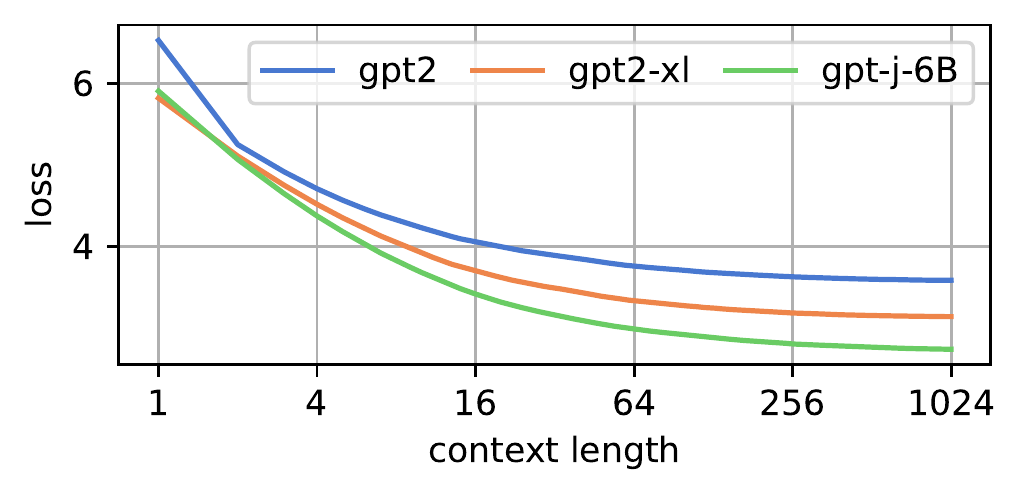}
    \caption{Mean LM losses by context length.}
    \label{fig:xent-ctx-models}
\end{figure}

\cref{fig:xent-ctx-models} shows the cross entropy losses (NLL means) across the whole validation dataset as a function of context length $c$.
As expected, larger models perform better than smaller ones, which is traditionally explained by their larger capacity.
A less common observation we can make thanks to this detailed representation is that the gains in performance come mostly from relatively short contexts (8--256 tokens); this is consistent with prior works
\citep{sun-etal-2021-long,press-etal-2021-shortformer}
which found that very long contexts bring only minimal improvement (though these focused on specific \emph{long-range} architectures and on contexts beyond the range we investigate here).

\begin{figure}
    \centering
    \includegraphics[width=0.95\linewidth,trim={0 0.25cm 0 0.2cm},clip]{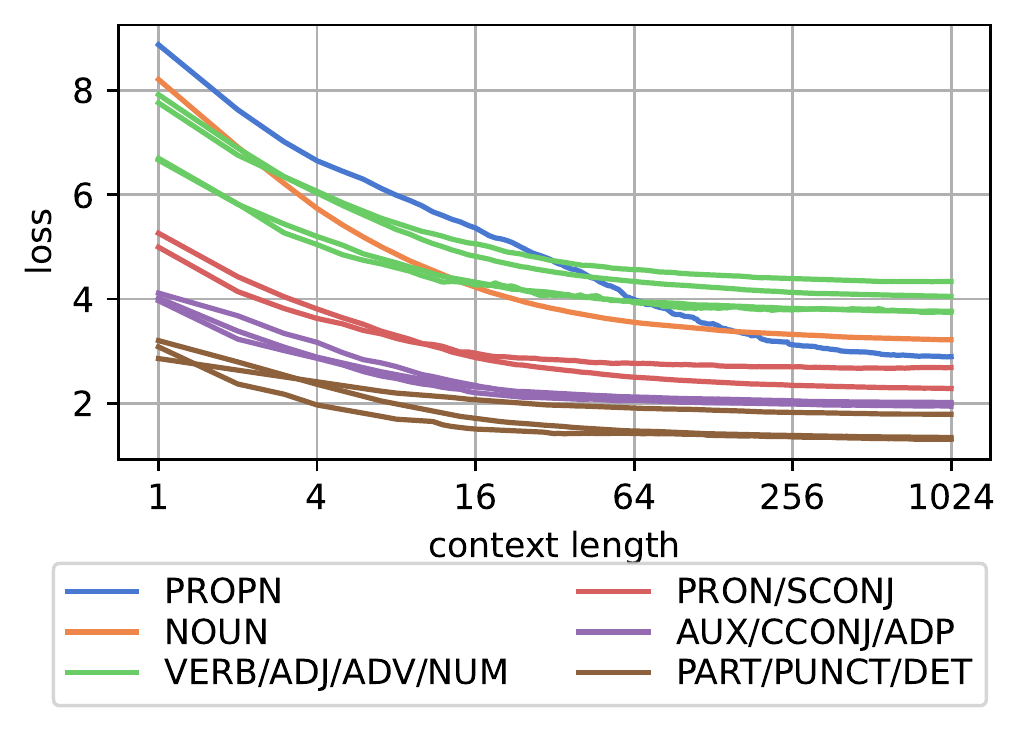}
    \caption{Mean GPT-J loss by context length and part-of-speech (POS) tag of the target token. Only POS tags with at least \num{100} occurrences in the dataset are included. The tags are grouped (arbitrarily) for clarity.}
    \label{fig:xent-ctx-pos}
\end{figure}

In \cref{fig:xent-ctx-pos}, we display the same information (loss by context length) broken down by part-of-speech (POS) tags, for GPT-J only.
For most POS tags, the behavior is similar to what we observed in \cref{fig:xent-ctx-models} and the loss appears to stabilize around context lengths \numrange{16}{64}.
However, we see a distinct behaviour for proper nouns (PROPN), which are the hardest-to-predict category for short contexts, but whose loss improves steadily with increasing $c$, surpassing that of regular nouns (NOUN) at $c=162$ and continuing to improve beyond that point. %

\begin{figure*}
    \centering
    \includegraphics[scale=.68,trim={0.25cm 0.25cm 0.25cm 0},clip]{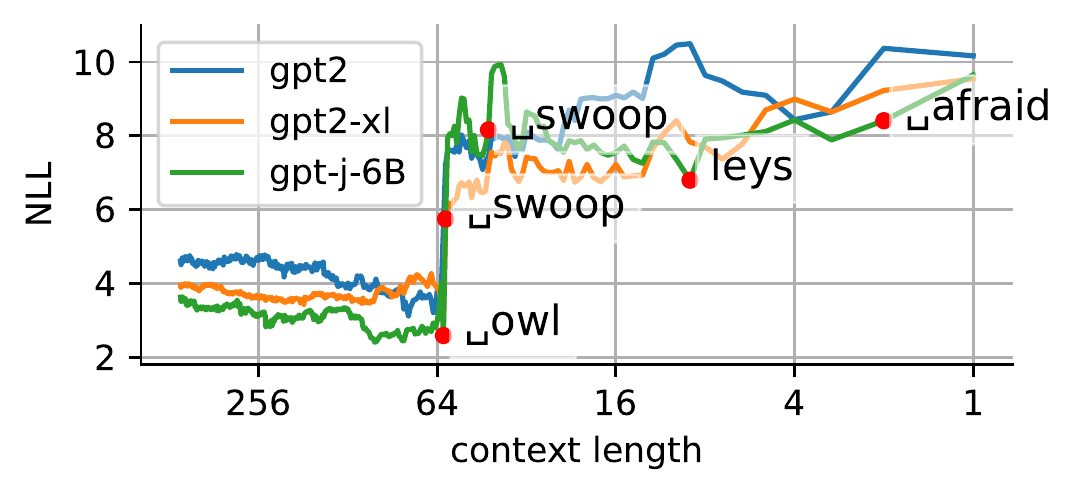}\quad
    \includegraphics[scale=.68,trim={0.25cm 0.25cm 0.25cm 0},clip]{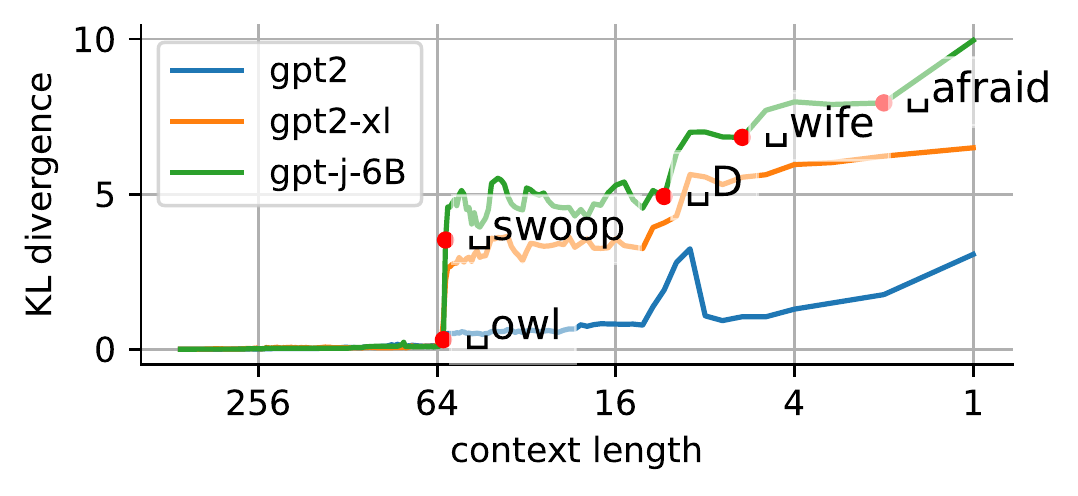}
    \caption{NLL (left) and KL divergence (right) as a function of context length for a selected example: ``\textelp{} mortally afraid of \textbf{birds}'' (same as in \cref{fig:ui-screenshot}).
    The $x$ axis is reversed for visual correspondence with the left-hand context.
    The 5 context tokens causing the largest drops in each metric for GPT-J are marked by red dots.
    }
    \label{fig:ex-birds-small}
\end{figure*}

\subsection{Per-token losses by context length}

We have also examined token-level losses, as well as the KL divergence metric (see \cref{sec:metrics});
an example plot is shown in \cref{fig:ex-birds-small} and more are found in \cref{sec:appendix-tokenwise-plots}.
In general, we observe that the values tend to change gradually with $c$; large differences are sparse, especially for large $c$, and can often be attributed to important pieces of information appearing in the context (e.g.\ ``owl'' and ``swoop'' in the context of ``birds'' in \cref{fig:ex-birds-small}).
This justifies our use of these differences as importance scores.

\subsection{Differential importance scores}
To facilitate the exploration of $\Delta$-scores from \cref{sec:delta-scores}, we have created an interactive web demo,\cref{foot:demo-url}
which allows visualizing the scores for any of the 3 models on the validation set as shown in \cref{fig:ui-screenshot}.

\begin{figure}[t]
    \centering
    \includegraphics[width=0.9\linewidth,trim={0.25cm 0.25cm 0.25cm 0.2cm},clip]{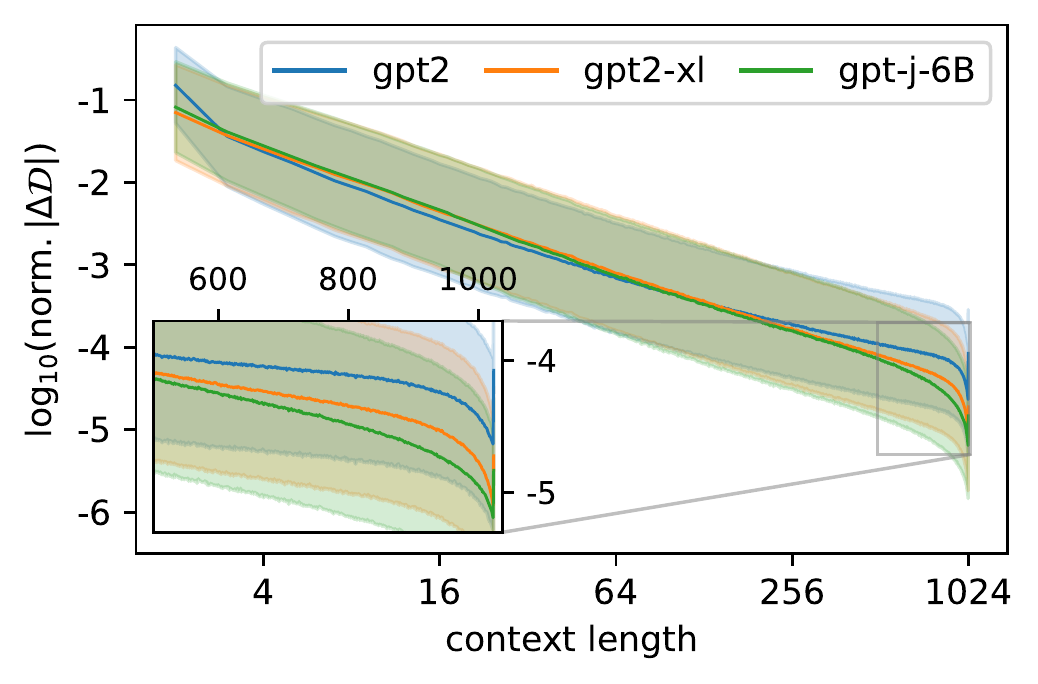}
    \caption{Normalized $\Delta$-score log-magnitude (mean and std.\ dev.)\ by context length and by model. Only positions $n\geq 1024$ are included.}
    \label{fig:ctx-len-imp}
\end{figure}

\begin{figure}[t]
    \centering
    \includegraphics[width=0.9\linewidth,trim={0.25cm 0.25cm 0.25cm 0.2cm},clip]{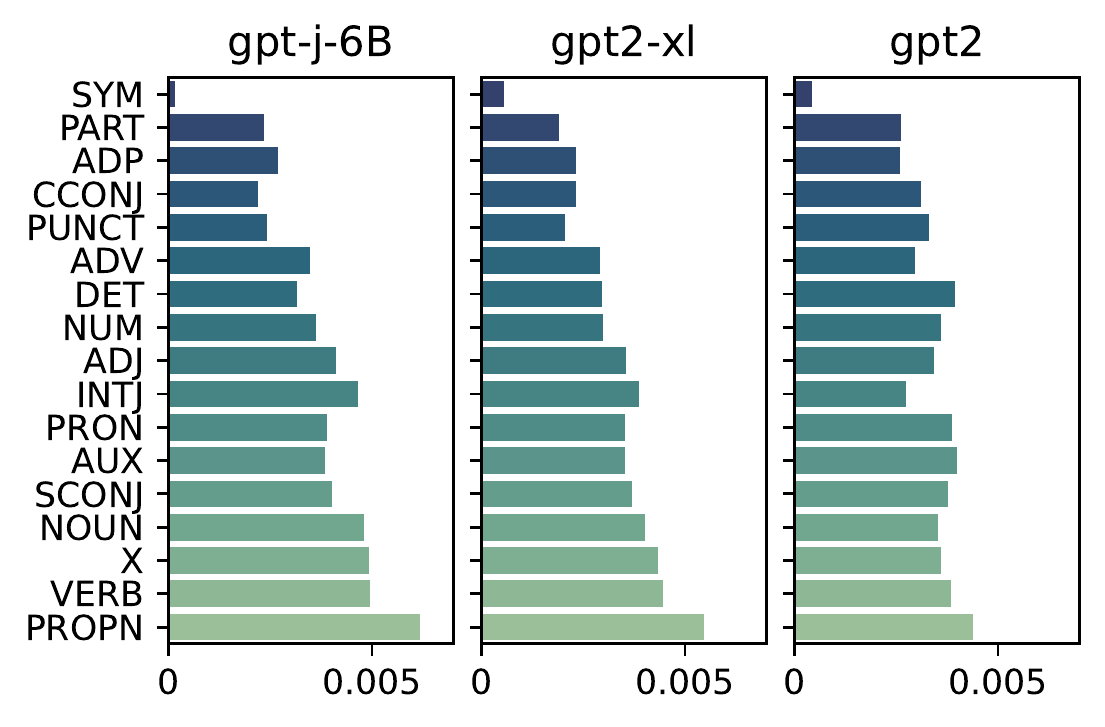}
    \caption{Mean $\Delta$-score by POS tag of the context token and by model.}
    \label{fig:upos-scores-single}
\end{figure}

In \cref{fig:ctx-len-imp}, we display the magnitudes of the $\Delta$-scores~-- normalized for each position to sum up to~1 across all context lengths~-- as a function of context length.
The plot suggests a power-law-like inverse relationship where increasing context length proportionally reduces the $\Delta$-score magnitude on average.
We interpret this as far-away tokens being less likely to carry information not already covered by shorter contexts.
Long contexts (see inset in \cref{fig:ctx-len-imp}) bear less importance
for larger models than for smaller ones, perhaps because the additional capacity allows relying more on shorter contexts.

In \cref{fig:upos-scores-single}, we also display the mean importance score received by each POS category, by model.
We can see that proper nouns (PROPN) are substantially more informative than other categories (which is in line with the observations in the previous section), but less so for the smallest model. This could mean e.g.\ that larger models are better at memorizing named entities from training data and using them to identify the topic of the document, or simply at copying them from distant context as observed in \citep{sun-etal-2021-long}.

\section{Limitations and future directions}

\paragraph{Experiments.}
We acknowledge the limited scope of our experiments, including only 8 (closed-domain) documents, 3 models and a single language.
This is largely due to the limited availability of suitable large LMs and their high computational cost.
Still, we believe that our experiments are valuable as a case study that already clearly showcases some interesting features of our methodology.

\paragraph{Computational cost.}
While we have de\-mons\-trated an efficient strategy to obtain predictions for all tokens at all possible context lengths, it still requires running the model $N$ times for a document of length $N$.%

For a $k$-fold reduction in computational cost, the technique may be modified to use a sliding window with stride $k>1$ (instead of $k=1$ as proposed above). %
See \cref{sec:appendix-strided} for details.

\paragraph{Choice of metrics.}
The proposed methodology allows investigating how any given metric is impacted by context, yet our study is limited to NLL loss and the proposed KL divergence metric (the latter for defining importance scores). These may not be optimal for every purpose, and other choices should be explored depending on the application.
For example, to study sequences \emph{generated} (sampled) from a LM, one might want to define importance scores using a metric that does depend on the generated token, e.g.\ its NLL loss or its ranking among all candidates.
(Indeed, our web demo also supports $\Delta$-scores defined using NLL loss values.)

\section{Conclusion and future directions}
We have presented \emph{context length probing}, a novel causal LM explanation technique based on tracking the predictions of the LM as a function of context length, and enabling the assignment of \emph{differential importance scores} (\emph{$\Delta$-scores}).
While it has some advantages over existing techniques, it answers different questions, and should thus be thought of as complementary rather than a substitute.

A particularly interesting feature of our $\Delta$-scores is their apparent potential for discovering \emph{long-range dependencies} (LRDs) (as they are expected to highlight information not already covered by shorter contexts, unlike e.g.\ attention maps).

Remarkably, our analysis suggests a power-law-like inverse relationship between context length and importance score, seemingly questioning the importance of LRDs in language modeling. While LRDs clearly appear crucial for applications such as long-form text generation, their importance may not be strongly reflected by LM performance metrics like cross entropy or perplexity.
We thus believe that there is an opportunity for more specialized benchmarks of LRD modeling capabilities of different models, such as that of \citet{sun-etal-2022-chapterbreak}, for example.
These should further elucidate questions like to what extent improvements in LM performance are due to better LRD modeling, how LRDs are handled by various Transformer variants \citep[e.g.][]{KitaevKL20,katharopoulosTransformersAreRNNs,choromanskiRethinkingAttentionPerformers2020a,Press22Alibi}, or what their importance is for different tasks.

\FloatBarrier

\iffinalcopy
\section*{Acknowledgments}
This work was supported by the LabEx NUMEV (ANR-10-LABX-0020) within the I-Site MUSE (ANR-16-IDEX-0006).
The authors are grateful to the OPAL infrastructure from Université Côte d'Azur for providing resources and support.
\fi

\bibliography{anthology,custom}

\clearpage
\appendix
\onecolumn

\section{Context length probing}
\label{sec:appendix-context-probing}

\begin{figure}
    \centering
    \includegraphics[width=0.6\linewidth]{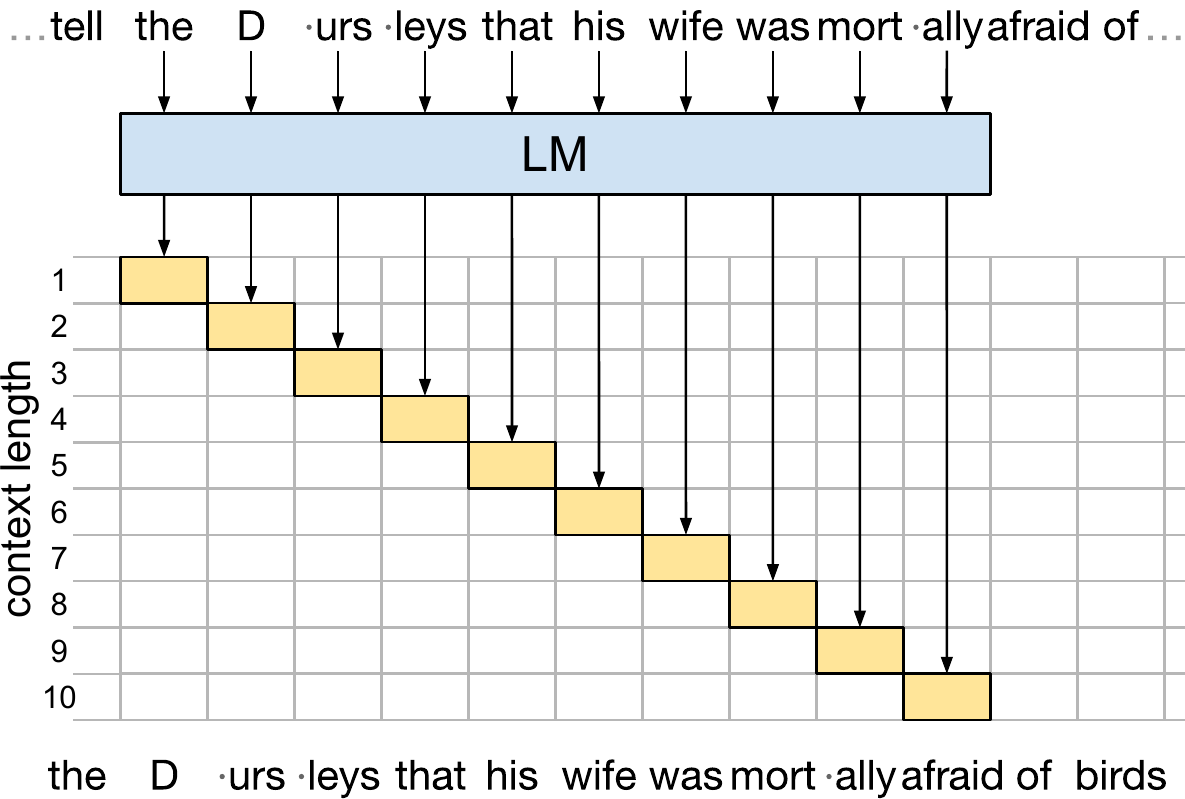}
    \caption{A step of context length probing with $c_\text{max}=10$. The input tokens are shown at the top, the target tokens at the bottom. When the LM is run on a segment of the document, the effective context length for each target token is equal to its offset from the beginning of the segment, e.g.\ the context for predicting ``\textvisiblespace D'' is ``\textvisiblespace the'' ($c=1$), the context for ``urs'' is ``\textvisiblespace the\textvisiblespace D'' ($c=2$), etc.}
    \label{fig:lm-eval-example}
\end{figure}

\cref{fig:lm-eval-example} illustrates a step of context length probing.
We wish to obtain the tensor $\bm{P}$ from \cref{eq:p}, understood as a table where each cell contains the predictions (next-token logits) for a given position in the text and a given context length.
By running our LM on a segment of the text, we get predictions such that for the $n$-th token in the segment, the effective context length is equal to $n$, which corresponds to a diagonal in the table. We can thus fill in the whole table by running the LM on all segments of length $c_\text{max}$ (plus trailing segments of lengths $c_\text{max}-1,\ldots,1$).

Notice that this process is somewhat similar to (naïvely) running the LM in generation mode, except that at each step, the leading token is removed, preventing the use of caching to speed up the computation.

In practice, it is not necessary to explicitly construct the tensor $\bm{P}$.
Indeed, we find it more efficient to instead store the raw logits obtained by running the model on all the segments, then do the necessary index arithmetics when computing the metrics.

\subsection{Strided context length probing}
\label{sec:appendix-strided}
For a $k$-fold reduction in computational cost, we may instead use a sliding window with a stride $k>1$, i.e.\ run the model only on segments starting at positions $k\left(n-1\right)+1$ for all $n\in\left\{1,\ldots,\left\lceil N/k\right\rceil\right\}$, rather than all positions.
This way, for a target token $x_{n+1}$, we obtain the predictions $p\mleft(x_{n+1}\;\middle|\;x_{n-c+1},\ldots,x_n\mright)$ only for such context lengths $c$ that $c\bmod k=n$. In other words, predictions with context length $c$ are only available for tokens $x_{c+1},x_{c+k+1},x_{c+2k+1},\ldots$.
Consequently:
\begin{itemize}
    \item Overall, we still cover all context lengths $1,\ldots,c_\text{max}$, allowing us to perform aggregate analyses like the ones in \cref{sec:loss-by-context-length}. 
    \item When analyzing the predictions for a specific target token in a document (e.g.\ to compute $\Delta$-scores), context tokens come in blocks of length $k$. Visualizations like the ones in \cref{fig:ui-screenshot,fig:ex-birds-small} are still possible for all target tokens, but become less detailed, grouping every $k$ context tokens together.
    \item Computation time, as well as the space needed to store the predictions, is reduced by a factor of $k$.
\end{itemize}

\section{Technical details}
\label{sec:appendix-computation-details}

\paragraph{Data.}
The LinES treebank is licensed under Creative Commons BY-NC-SA 4.0.
We concatenated all tokens from each of the documents from the treebank, then re-tokenized them using the GPT-2 tokenizer. We mapped the original (UD) POS tags to the GPT-tokenized dataset in such a way that every GPT token is assigned the POS tag of the first UD token it overlaps with.

\paragraph{Models.} We used the models \href{https://huggingface.co/EleutherAI/gpt-j-6B}{\texttt{EleutherAI/gpt-j-6B}} (Apache 2.0 license), and \href{https://huggingface.co/gpt2-xl}{\texttt{gpt2-xl}} and \href{https://huggingface.co/gpt2}{\texttt{gpt2}} (MIT license), all from \href{https://huggingface.co/}{\texttt{huggingface.co}}.

\paragraph{Computation.} We parallelized the inference over \num{500} jobs on a compute cluster,\footnote{\anon{Nef}, the cluster computing infrastructure of \anon{Inria Sophia Antipolis Méditerranée; see \url{https://wiki.inria.fr/ClustersSophia}}} each running on 8 CPU cores with at least \SI{8}{\giga\byte} of RAM per core, with a batch size of \num{16}.
Each job took about \SIrange{10}{20}{\minute} for GPT-2 and \SIrange{30}{60}{\minute} for GPT-J.
Additionally, computing the metrics from the logits (which take up \SI{2}{\tera\byte} of disk space in \texttt{float16}) took between \num{2} and \SI{4}{\hour} per model on a single machine with \num{32} CPU cores.
The total computing time was \num{318} core-days, including debugging and discarded runs.

\section{Additional plots}
\label{sec:appendix-plots}

\subsection{Token-wise metrics as a function of context length}
\label{sec:appendix-tokenwise-plots}
\begin{figure*}
    \centering
    \begin{subfigure}{\linewidth}
    \centering
    \includegraphics[scale=0.65]{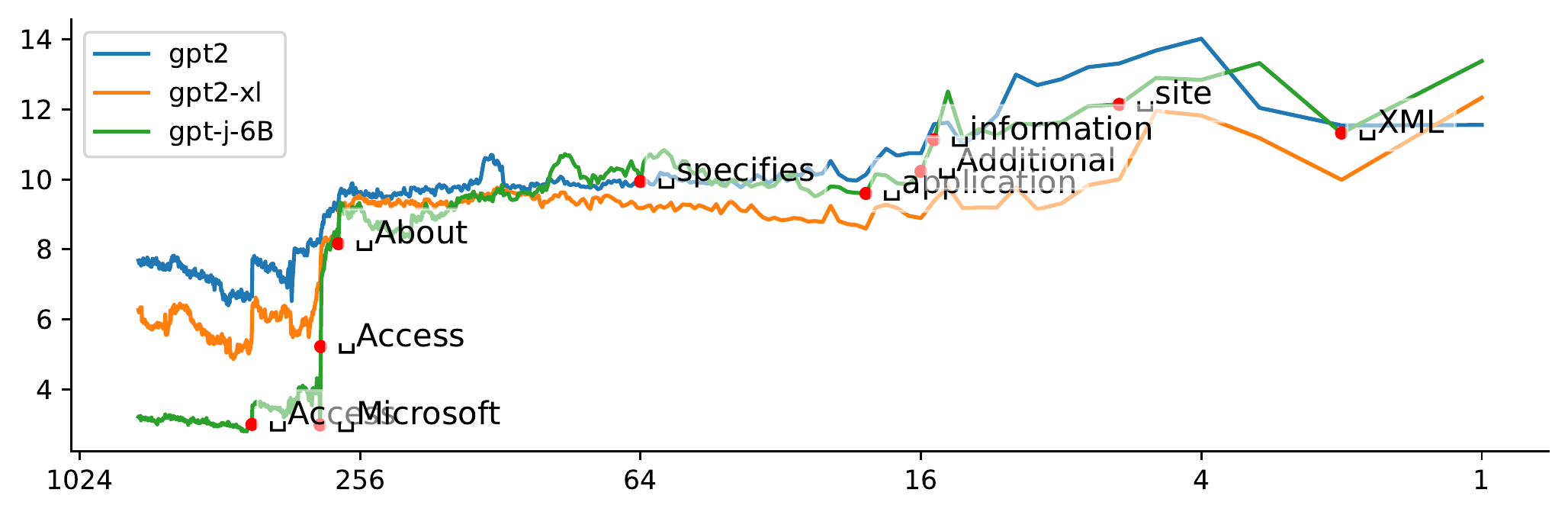}
    \caption{\dots and attribute means (and thus how the data between them will look in a browser), XML uses the tags only to delimit pieces of data, and leaves the interpretation of the data completely to the application that reads it. Additional information about XML can be found on the web site. About importing XML data\textbf{ Access}}
    \end{subfigure}
    \begin{subfigure}{\linewidth}
    \centering
    \includegraphics[scale=0.65]{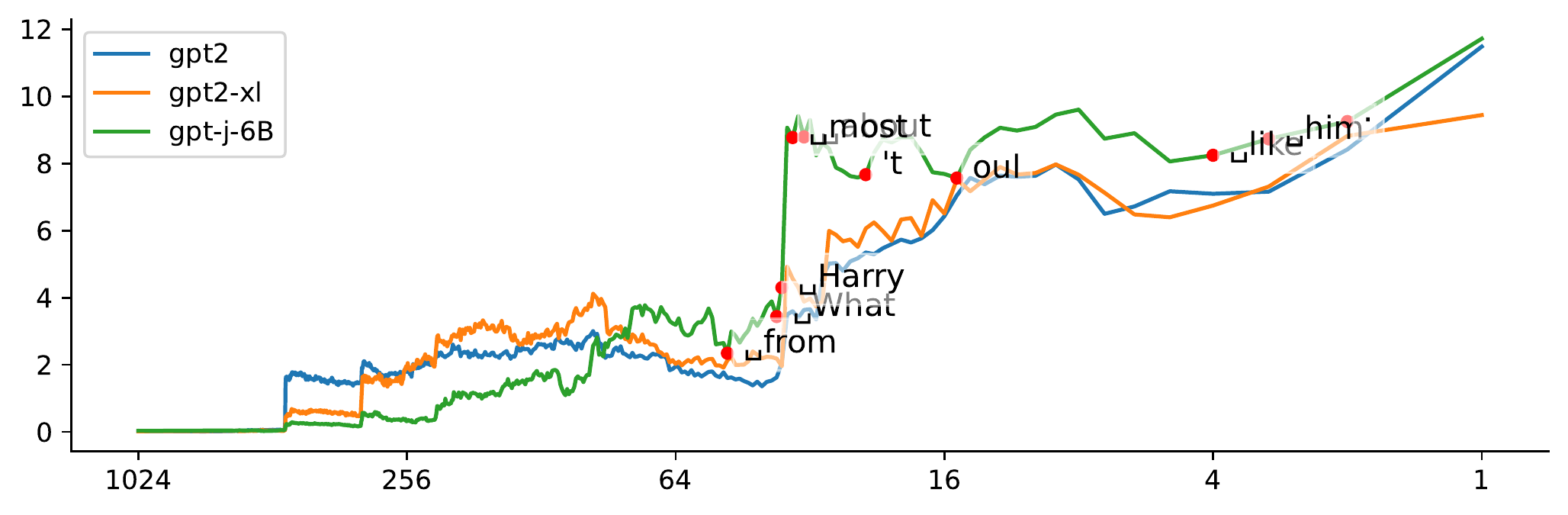}
    \caption{\dots he felt things were getting too quiet, and small explosions from Fred and George's bedroom were considered perfectly normal. What Harry found most unusual about life at Ron's, however, wasn't the talking mirror or the clanking ghoul: it was the fact that everybody there seemed to like him. Mrs\textbf{ Weasley}}
    \end{subfigure}
    \begin{subfigure}{\linewidth}
    \centering
    \includegraphics[scale=0.65]{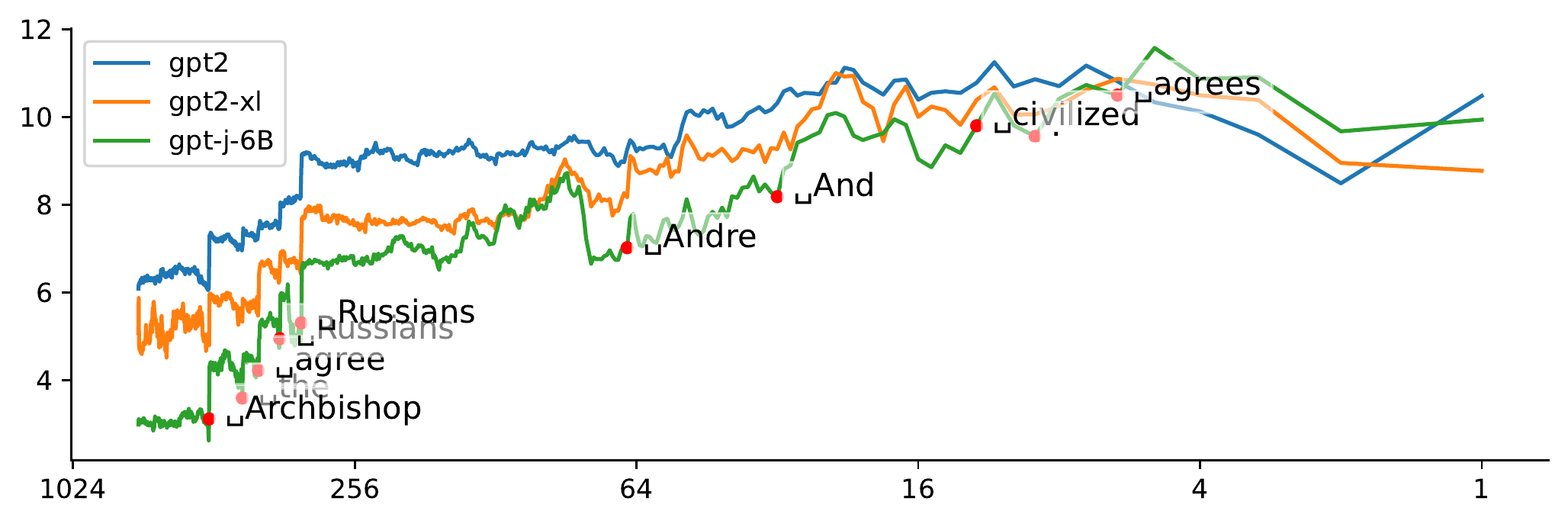}
    \caption{\dots is a great difference between Napoleon the Emperor and Napoleon the private person. There are raisons d'etat and there are private crimes. And the talk goes on. What is still being perpetuated in all civilized discussion is the ritual of civilized discussion itself. Tatu agrees with the Archbishop about the\textbf{ Russians}}
    \end{subfigure}
    \caption{NLL losses ($y$ axis) for 3 selected target tokens as a function of context length ($x$ axis). Below each plot, the target token is displayed in bold, along with a context of 60 tokens. The $x$ axis is reversed to correspond visually to left-hand context. The red dots show the 10 tokens that cause the largest drops in GPT-J cross entropy when added to the context.}
    \label{fig:xent-example-plots}
\end{figure*}

\begin{figure*}
    \centering
    \begin{subfigure}{\linewidth}
    \centering
    \includegraphics[scale=0.65]{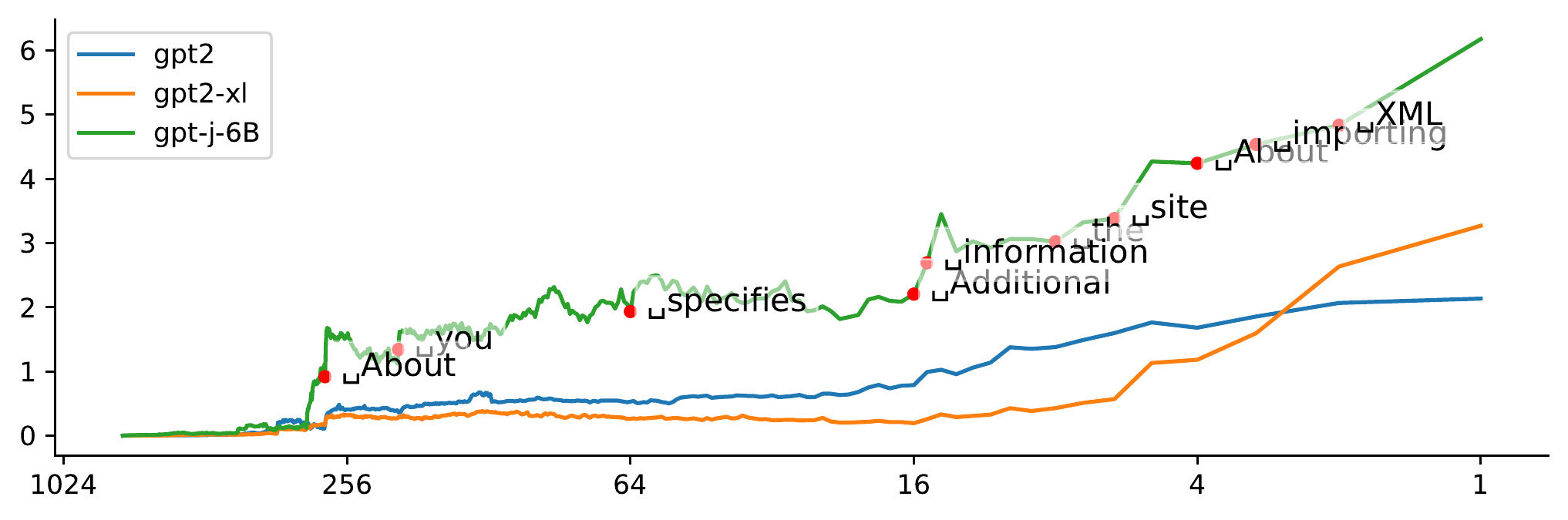}
    \caption{\dots and attribute means (and thus how the data between them will look in a browser), XML uses the tags only to delimit pieces of data, and leaves the interpretation of the data completely to the application that reads it. Additional information about XML can be found on the web site. About importing XML data\textbf{ Access}}
    \end{subfigure}
    \begin{subfigure}{\linewidth}
    \centering
    \includegraphics[scale=0.65]{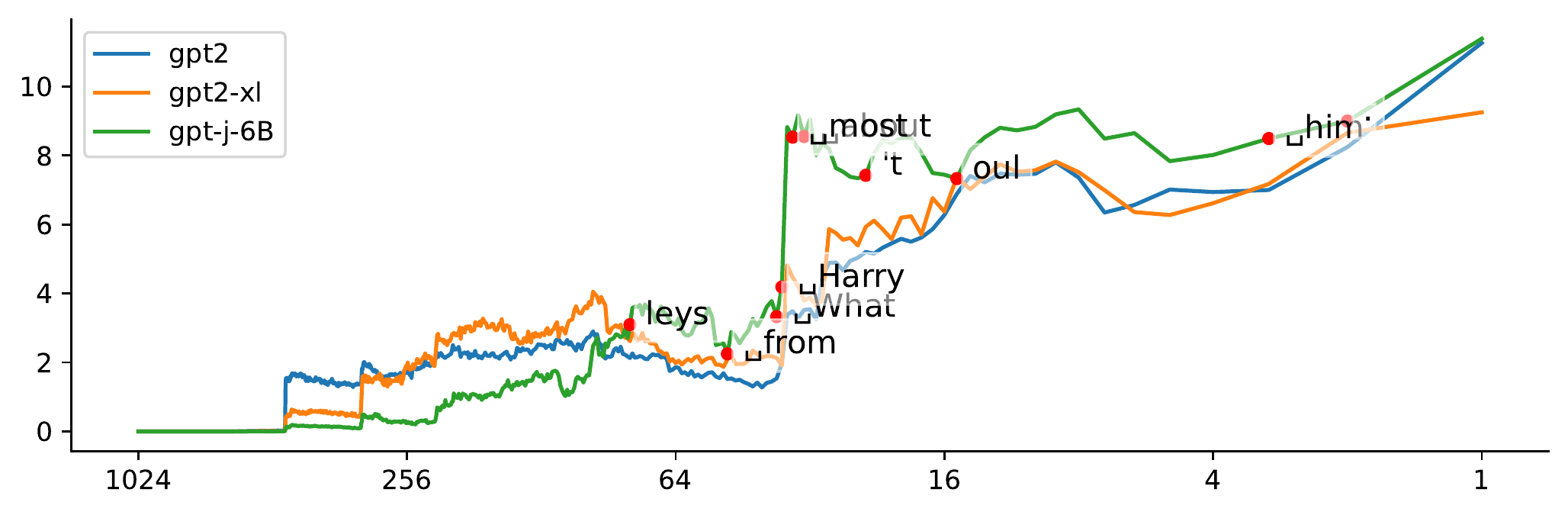}
    \caption{\dots he felt things were getting too quiet, and small explosions from Fred and George's bedroom were considered perfectly normal. What Harry found most unusual about life at Ron's, however, wasn't the talking mirror or the clanking ghoul: it was the fact that everybody there seemed to like him. Mrs\textbf{ Weasley}}
    \end{subfigure}
    \begin{subfigure}{\linewidth}
    \centering
    \includegraphics[scale=0.65]{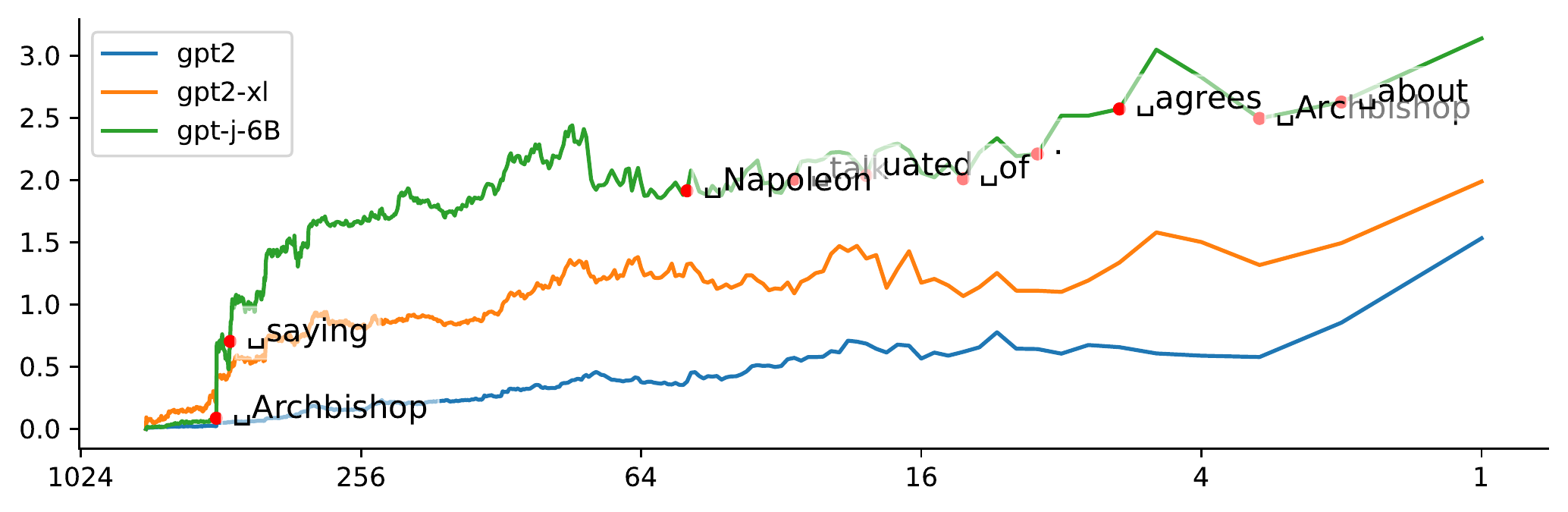}
    \caption{\dots is a great difference between Napoleon the Emperor and Napoleon the private person. There are raisons d'etat and there are private crimes. And the talk goes on. What is still being perpetuated in all civilized discussion is the ritual of civilized discussion itself. Tatu agrees with the Archbishop about the\textbf{ Russians}}
    \end{subfigure}
    \caption{KL divergences ($y$ axis) from \cref{eq:kl-div-metric} for 3 selected target tokens as a function of context length ($x$ axis). Below each plot, the target token is displayed in bold, along with a context of 60 tokens. The $x$ axis is reversed to correspond visually to left-hand context. The red dots show the 10 tokens that cause the largest drops in the metric (for GPT-J) when added to the context.}
    \label{fig:kl-example-plots}
\end{figure*}

\cref{fig:xent-example-plots,fig:kl-example-plots} show NLL and KL divergence \eqref{eq:kl-div-metric}, respectively, as a function of context length, for selected target tokens (proper nouns) from the validation set.

\end{document}